\documentclass{bmvc2k}

\title{Learning Visual Actions Using Multiple Verb-Only Labels}

\addauthor{Michael Wray}{michael.wray@bristol.ac.uk}{1}
\addauthor{Dima Damen}{dima.damen@bristol.ac.uk}{1}

\addinstitution{
 University of Bristol, UK
}

\runninghead{Wray, Damen}{Learning Visual Actions using Multiple Verb-Only Labels}

\usepackage{booktabs}
\usepackage{tabularx}
\usepackage{bm}
\usepackage{color}
\usepackage{amsfonts}

\def\eg{\emph{e.g}\bmvaOneDot}

\def\etal{\emph{et al}\bmvaOneDot}
\def\ie{\emph{i.e}\bmvaOneDot}

\newcommand{\dima}[1]{\textcolor{black}{#1}}

\newcommand{\qword}[1]{\emph{``#1''}}
\newcommand{\labelA}{Soft Assigned Multi-Verb Label}
\newcommand{\pphi}[1]{{$\phi_{#1}$}}

\newcommand{\myparagraph}[1]{\vspace*{-6pt} \paragraph{#1.}}

\begin{document}

\maketitle

\begin{abstract}
This work introduces verb-only representations for both recognition and retrieval of visual actions\dima{, in video}.
Current methods neglect legitimate semantic ambiguities between verbs, instead choosing unambiguous subsets of verbs along with objects to disambiguate the actions.
\dima{We instead propose multiple verb-only labels, which we learn through hard or soft assignment as a regression. This enables learning}
a much larger vocabulary of verbs, including contextual overlaps of these verbs.
We collect multi-verb annotations for three action \dima{video} datasets and evaluate the verb-only labelling representations for action recognition and cross-modal retrieval (video-to-text and text-to-video). We demonstrate that multi-label verb-only representations outperform conventional single verb labels.
We also explore other benefits of a multi-verb representation including cross-dataset retrieval and verb type (manner and result verb types) retrieval.
\end{abstract}

%%%%%%%%%%%%%%%%%%%%%%%%%%%%%%%%%%%%%%%%%%%%%%%%%%%%%%%%%%%%%%%%%%%%%%%%%%
\section{Introduction} \label{sec:intro}

\begin{figure}[t]
    \centering
    \includegraphics[width=\textwidth]{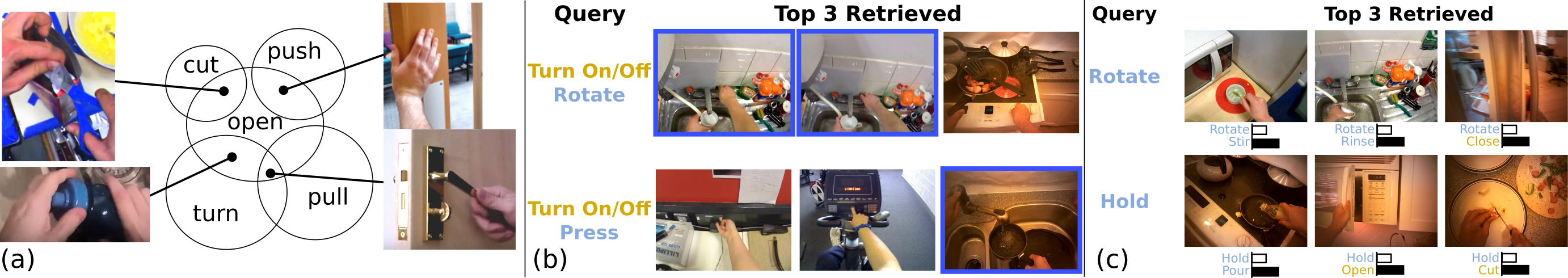}
    \vspace{-10pt}
    \caption{(a) Open can be performed in multiple ways leading to complex overlaps in the space of verbs. Using multiple verb-only representations allows: (b) Retrieval that distinguishes between different manners of performing the action.  Actions of turning on/off by \qword{rotating} are separated from those of turning on/off by \qword{pressing}; 
    (c) Verbs such as \qword{rotate} and \qword{hold} can be learned via context from multiple actions.}
    \label{fig:main}
\end{figure}

With the rising popularity of deep learning, action recognition datasets are growing in 
number of videos and scope~\cite{Damen2018EPICKITCHENS, goyal2017_SomethingSomethingVideo, kay2017kinetics, sigurdsson2016hollywood, li2018eye, gu2018ava}, leading to an increasingly large vocabulary required to label them.
This introduces challenges in labelling, particularly as more than one verb is relevant to each action.
For example, consider the common action of opening a door and the verbs that could be used to describe it.
\qword{Open} is a natural choice, with \qword{pull} being used to specify the direction as well as \qword{grab}, \qword{hold} and \qword{turn} characterising the method of turning the handle.
Any action can thus be richly described by a number of different verbs.
This is contrasted with current approaches which use a reduced set of semantically distinct verb labels, combined with the object being interacted with (e.g.~\qword{open-door}, \qword{close-jar}).
Sigurdsson~\etal~\cite{sigurdsson2017actions} show that using only single verb labels, without any accompanying nouns, leads to increased confusion among human annotators. Conversely, verb-noun labels
\dima{cause} actions to be tied to instances of objects, \ie opening a fridge is the same interaction as opening a cupboard\dima{. Additionally, using a single verb} doesn't explore the complex overlapping space of verbs in visual actions\dima{. In }  Fig.~\ref{fig:main}(a)\dima{, four `open' actions are shown, where the verb `open' overlaps with four different verbs: `push' and `pull' for two doors, `turn' when opening a bottle and `cut' when opening a bag.}

In this paper, we deviate from previous works which use a reduced set of verb and noun pairs~\cite{Damen2014a, Damen2018EPICKITCHENS, taralova2011source, Fathi2012, kay2017kinetics, Kuehne11, sigurdsson2016hollywood, soomro2012ucf101}, and instead propose using a representation of multiple verbs to describe the action.
The combination of multiple, individually ambiguous verbs allow for an object-agnostic labelling approach,  which we define using an assignment score for each verb. 
We investigate using both hard and soft assignment, and evaluate the representations on two tasks: action recognition and cross-modal retrieval (\ie video-to-text and text-to-video retrieval).
Our results show that hard-assignment is suitable for recognition while soft-assignment improves cross-modal retrieval.

A multi-verb representation presents a number of advantages over a verb-noun representation.
Figure~\ref{fig:main}(b) demonstrates this by querying the multi-verb representation with \qword{turn-off}
combined with one other verb (\qword{rotate} vs. \qword{press}).
Our representation can differentiate between a tap turned off by rotating (first row in blue) from one turned off by pressing (second row in blue). \dima{Both single verb representations (i.e. `turn-off') and verb-noun representations (i.e. `turn-off tap') cannot learn to distinguish these different interactions.}
In Fig~\ref{fig:main}(c), verbs such as \qword{rotate} and \qword{hold}, describing parts of the action, can also be learned via context of co-occurrence with other verbs.

To the best of the author's knowledge, (\textbf{our contributions}): (i)
this paper proposes to address the inherent ambiguity of verbs, by using multiple verb-only labels to represent actions, keeping these representations object-agnostic.
(ii) We annotate these multi-verb representations for three public \dima{video} datasets
(BEOID~\cite{wray2016sembed}, CMU~\cite{taralova2011source} and GTEA Gaze+~\cite{Fathi2012}).
(iii) We evaluate the proposed representation for two tasks: action recognition and action retrieval, and show that hard assignment is more suited for recognition tasks while soft assignment assists retrieval tasks, including for cross-dataset retrieval.

%%%%%%%%%%%%%%%%%%%%%%%%%%%%%%%%%%%%%%%%%%%%%%%%%%%%%%%%%%%%%%%%%%%%%%%%%%
\section{Related Work} \label{sec:related_work}

\myparagraph{Action Recognition in Videos}
Video Action Recognition datasets are commonly annotated with a reduced set of semantically distinct verb labels ~\cite{Damen2018EPICKITCHENS,Damen2014a,de2008guide,Fathi2012,goyal2017_SomethingSomethingVideo,gu2018ava, sigurdsson2018actor,sigurdsson2016hollywood}.
Only in EPIC-Kitchens~\cite{Damen2018EPICKITCHENS}, verb labels \dima{are collected from narrations with} an open vocabulary leading to overlapping labels, which are then manually clustered into unambiguous classes.
Ambiguity and overlaps in verbs has been noted in~\cite{wray2016sembed,khamis2015walking}. 
\dima{Our prior work~\cite{wray2016sembed} uses} the verb hierarchy in WordNet~\cite{miller1995wordnet} synsets to reduce ambiguity.
\dima{We note} how annotators were confused, and often could not distinguish between the different verb meanings.
Khamis and Davis~\cite{khamis2015walking} used multi-verb labels in action recognition, on a small set of (10) verbs.
They jointly learn multi-label classification and label correlation, using a bi-linear approach, allowing an actor to be in a state of performing multiple actions such as \qword{walking} and \qword{talking}.
This work is the closest to ours in motivation, however their approach uses hard assignment of verbs, and does not address single-verb ambiguity, assuming each verb to be non-ambiguous.
Up to our knowledge, no other work has explored multi-label verb-only representations of actions in video.

\myparagraph{Action Recognition in Still Images}
Gella and Keller~\cite{gella17analysis} present an overview of datasets for action recognition from still images. Of these, four use a large (${>}50$) number of verbs as labels~\cite{chao2015benchmark,gella2016unsupervised,ronchi2015describing,yatskar2016situation}.
\mbox{ImSitu~\cite{yatskar2016situation}} uses single verb labels, however, they note ambiguities between verbs --- with each verb having an average of 3.55 different meanings. They report top-1 and top-5 accuracies to account for the multiple roles each verb plays.
The HICO dataset~\cite{chao2015benchmark} has multi-label verb-noun annotations for its 111 verbs, but the authors combine the labels into a powerset to avoid the overlaps between verbs. 
Two datasets, Verse~\cite{gella2016unsupervised} and COCO-a~\cite{ronchi2015describing}, use external semantic knowledge to disambiguate a verb's meaning.
For zero-shot action recognition of images, Zellers and Choi~\cite{zellers2017zero} use verbs in addition to attributes such as global motion, word embeddings, and dictionary definitions to solve both text-to-attributes and image-to-verb tasks.
While their labels are coarse grained (and don't describe object interactions) they demonstrate the benefit of using a verb-only representation.

Although works are starting to use a larger number of verbs, the class overlaps are largely ignored through the use of clustering or assumption of hard boundaries.
In this work we 
learn a multi-verb representation acknowledging these issues.

\myparagraph{Action Retrieval}
Distinct from recognition, cross-modal retrieval approaches have been proposed for visual actions both in images~\cite{gordo2017beyond,RadenovicECCV16CNN,zhang2016zero} and videos~\cite{dong2018dual,guadarrama2013youtube2text,xu2015jointly}.
These works focus on instance retrieval, \ie given a caption can the corresponding video/image 
be retrieved and vice versa.
This is different from our attempt to retrieve similar actions rather than only the corresponding video/caption.
Only Hahn~\etal~\cite{hahn2018action} train an embedding space for videos and verbs only, using word2vec as the target space.
They use verbs from UCF101~\cite{soomro2012ucf101} and HMDB51~\cite{Kuehne11} in addition to verb-noun classes from Kinetics~\cite{kay2017kinetics}.
These are coarser actions (\eg \qword{diving} vs. \qword{running}) and as such have little overlap allowing the target space to perform well on unseen actions.

As far as we are aware, ours is the first work to use the same representation for both action recognition and action retrieval.

\myparagraph{Video Captioning}
Semantics and annotations have also been studied by the recent surge in generating captions for images~\cite{mao2014deep, devlin2015language, anne2016deep, aneja2018convolutional, anderson2018bottom} and videos~\cite{venugopalan2014translating, yao2015describing, yu2016video, zhou2018end, wang2018video, wang2018reconstruction}.
These works assess success by the acceptability of the generated caption.
We differ in our focus on learning a mapping function between an input video and verbs in the label set, thus providing a richer understanding of the overlaps between verb labels.

%%%%%%%%%%%%%%%%%%%%%%%%%%%%%%%%%%%%%%%%%%%%%%%%%%%%%%%%%%%%%%%%%%%%%%%%%%

\vspace*{-12pt}
\section{Proposed Multi-Verb Representations} \label{sec:annotations}
\vspace*{-4pt}

In this section, we introduce types of verbs used to describe an action and note their absence from semantic knowledge bases
(Sec.~\ref{subsec:verb_meanings}).
We then define the single and multi-verb representations (Sec.~\ref{subsec:action_representations}).
Next, we describe the collection of the multiple verb annotations~(Sec.~\ref{subsec:annotation_proc_stat}) and detail our approach to learn the representations (Sec.~\ref{subsec:method}).

\vspace*{-6pt}
%-------------------------------------------------------------------------
\subsection{Verb Types} \label{subsec:verb_meanings}
\vspace*{-2pt}

From the earlier example of a door being opened, we highlight the different types of verbs that can be used to describe the same action, and their relationships.
Firstly, as with standard multi-label tasks, some verbs are related semantically; e.g. \qword{pull} and \qword{draw} are synonyms in this context.
Secondly, verbs describe different linguistic types of the action: \emph{Result verbs} describe the end state of the action, and \emph{manner verbs} detail the manner in which the action is carried out~\cite{alka1998building, gropen1991affectedness}.
Here, \qword{open} is the result verb, whilst \qword{pull} is a manner verb.
This distinction of verb types for visual actions has not been explored before for action recognition/retrieval.
Finally, there are verbs that describe the sub-action of using the handle, such as \qword{grab}, \qword{hold} and \qword{turn}. These are also result/manner verbs. 
We call such verbs \emph{supplementary} as they are used to describe the sub-action(s) that the actor needs to perform in order to complete the \dima{overall} action. Furthermore, they are highly dependent on context.

While these verb relationships can be explicitly stated, they are 
not available in lexical databases and are hard to discover from public corpora.
We illustrate this for the two commonly used sources of semantic knowledge: {W}ord{N}et~\cite{miller1995wordnet} and {W}ord2{V}ec~\cite{mikolov2013efficient}.
\begin{center}
    \footnotesize
  \begin{tabular}{ll|c|c}
   Verb Relationships & &\textbf{\hspace{8pt}WordNet\hspace{8pt}} &\textbf{\hspace{8pt}Word2Vec\hspace{8pt}}\\ \hline
   synonyms        &(e.g. \qword{pull}-\qword{draw})  & $\checkmark$  & $\times$ \\
   result-manner   &(e.g. \qword{open}-\qword{pull})  & $\times$      & $\times$ \\
   supplementary   &(e.g. \qword{open}-\qword{hold})  & $\times$      & $\times$ \\ 
  \end{tabular}
\end{center}

\myparagraph{{W}ord{N}et~\cite{miller1995wordnet}} {W}ord{N}et is a lexical English Database with each word assigned to one or more synsets, each representing a different meaning of the word.
{W}ord{N}et defines multiple relationships between synsets, including synonyms and hyponyms,
but does not capture the other two types of relationships (result/manner, supplementary).
Moreover, using WordNet requires prior knowledge of the verb's synset.

\myparagraph{{W}ord2{V}ec~\cite{mikolov2013efficient}} {W}ord2{V}ec embeds words in a vector space, where cosine distances can be computed.
These distances are learned based on co-occurrences of words in a given corpus.
For example, using Wikipedia, the embedded vector of the verb \qword{pull} has high similarity to the embedded vector of \qword{push} even though these actions are antonyms. 
Co-occurrences depend on the corpus used, and do not cover any of the relationships noted above.
(\qword{pull} and \qword{draw}, for example do not frequently co-occur in Wikipedia). 
Even using a relevant corpus, such as TACoS~\cite{rohrbach14gcpr},  \qword{push} is closest to \qword{crush}, \qword{rough} and \qword{rotate} as the three most semantically similar verbs.

As these relationships cannot be discovered from semantic knowledge bases, we opt to crowdsource the multi-verb labels (see Sec.~\ref{subsec:annotation_proc_stat}).

%-------------------------------------------------------------------------
\subsection{Representations of Visual Actions} \label{subsec:action_representations}

\begin{figure}[t]
  \begin{center}
     \includegraphics[width=1\textwidth]{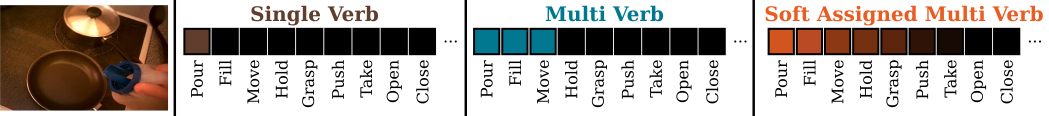}
  \end{center}
  \vspace{-20pt}
  \caption{Comparisons of different verb-only representations for \qword{Pour Oil} from GTEA+.}
  \label{fig:labelling_examples}
\end{figure}

\myparagraph{Single-Verb Representations} 
We start by defining the commonly-used single-label representation of visual actions. Each video $x_i \in X$ has a corresponding label $\bm{y_i} \in Y$ where $\bm{y_i}$ is a one-hot vector. In \textbf{Single-Verb Label (SV)}, this will be over verbs $V=\langle v_1, ..., v_N \rangle$ with $v_j$ representing the $j^{th}$ verb.
For example, in Fig.~\ref{fig:labelling_examples} the verbs \qword{pour} and \qword{fill} are both relevant but only one can be labelled, and more general verbs like \qword{move} won't be labelled either.
Additionally, we define a \textbf{Verb-Noun Label~(VN)} representation over $\langle v_j, n_k \rangle$ equivalent to the standard approach where a one-hot action vector is used. 

\myparagraph{Multi-Verb Representations}
We now propose two multi-verb representations.
First, \textbf{Multi-Verb Label (MV)}, $\bm{y_i}=\langle y_{i,j} \in \{0,1\} \rangle$ which is a binary vector over $V$ (\ie hard assignment).
Multiple verbs can be used to describe each video, e.g. \qword{pour} and \qword{fill}, allowing for manner and result verbs as well as semantically related verbs to be represented.
Hard label assignment though can be problematic for supplementary verbs.
Consider the verbs \qword{hold},\qword{grasp} in Fig.~\ref{fig:labelling_examples}.
These do not fully describe the object interaction yet cannot be considered irrelevant labels.
Including supplementary verbs in a multi-label representation would cause confusion between videos where the whole action is \qword{hold}, e.g. \qword{hold button}.

Alternatively, one can use a \textbf{\labelA~(SAMV)},~to increase the size of $V$ while accommodating supplementary verbs. 
Soft assignment assigns a numerical value to each verb, representing its relevance to the ongoing action.
In SAMV, each video $x_i$ will have a label vector $\bm{y_i}= \langle y_{i,j} \in [0,1] \rangle$ over $V$.
For two verb scores $y_{i,j} > y_{i,k} > 0$ when the verb $v_j$ is more relevant to the action $x_i$, while $v_k$ is still a valid/relevant verb.
This ranking makes this representation suitable for retrieval, and allows the
set of verbs, $V$, to be larger due to the restrictions of the previous representations being removed (Fig.~\ref{fig:labelling_examples}).

%-------------------------------------------------------------------------

\subsection{Annotation Procedure and Statistics} \label{subsec:annotation_proc_stat}

\begin{table}[b]
\footnotesize
    \centering
    \begin{tabularx}{\textwidth}{X>{\hsize=.005\hsize}XX}
    \toprule
    Manner & & Result \\
    \midrule
    carry, compress, drain, flip, fumble, grab, grasp, grip, hold, hold~down, hold~on, kick, let~go, lift, pedal, pick~up, point, position, pour, press, press~down, pull, pull~out, pull~up, push, put~down, release, rinse, reach, rotate, scoop, screw, shake, slide, spoon, spray, squeeze, stir, swipe, swirl, switch, take, tap, tip, touch, twist, turn & &
    activate, adjust, check, clean, close, crack, cut, distribute, divide, drive, dry, examine, fill, fill~up, find, input, insert, mix, move, open, peel, place, plug, plug~in, put, relax, remove, replace, return, scan, set~up, spread, start, step~on, switch~on, transfer, turn~off, turn~on, unlock, untangle, wash, wash~up, weaken \\
    \bottomrule
    \end{tabularx}
    \caption{Manual split of the original verbs.}
    \label{tab:manual_split}
\end{table}

To acquire the multi-verb representations we now describe the process of collecting multi-verb annotations for three action datasets: BEOID~\cite{Damen2014a} (732 video segments), CMU~\cite{taralova2011source} (404 video segments) and GTEA+~\cite{Fathi2012} (1001 video segments).
All three datasets are shot with a head mounted camera and include kitchen scenes.
First, we constructed the list of verbs, $V$, that annotators could choose from, by combining the unique verbs from all available annotations of the datasets
\cite{Fathi2012,de2008guide,Damen2014a,wray2016sembed}, giving a list of $90$ verbs - we exclude \{rest, read, walk\} to focus on object interactions.
\dima{Table~\ref{tab:manual_split} includes the manual split of verbs between manner and result.
We found that the datasets included an almost even spread of manner and result verbs with 47 and 43 respectively.}
We then chose one video per action, and asked 
30-50 annotators from Amazon Mechanical Turk (2,939 total) to select all verbs which apply. We normalise the responses by the number of annotators, so that for each verb, the score lies in the range of $[0,1]$. 

\begin{figure}[t]
  \begin{center}
  	\includegraphics[width=1\textwidth]{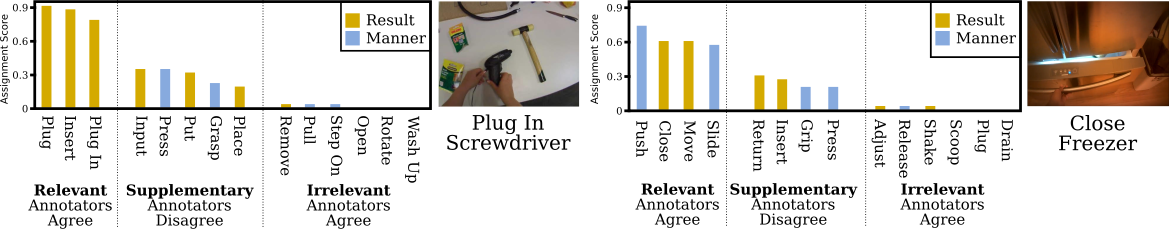}
  \end{center}
  \vspace{-12pt}
  \caption{
  When annotators agree, the verbs are highly relevant or certainly irrelevant. When annotators disagree, verbs are supplementary, for both \textit{result} and \textit{manner} verb types.} 
  \label{fig:verb_dist_all}
 \end{figure}
 \begin{figure}[t]
  \begin{center}
     \includegraphics[width=1\textwidth]{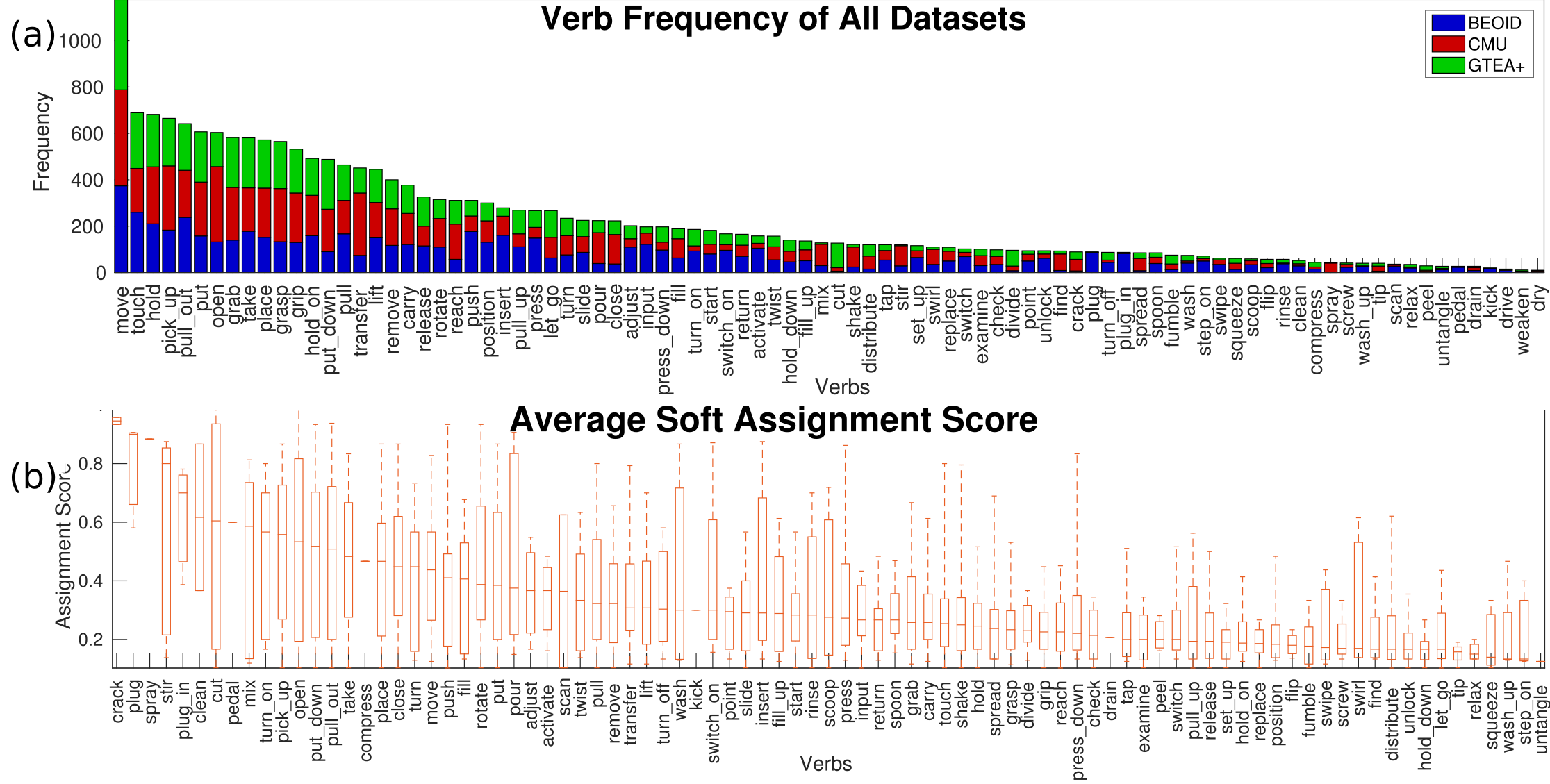}
  \end{center}
  \vspace{-20pt}
  \caption{\textbf{Annotation Statistics}. Figure best viewed in colour - (a) Number of annotations per verb. (b) Average soft assignment score per verb when chosen.}
  \label{fig:annotation_statistics}
\end{figure}

Figure~\ref{fig:verb_dist_all} shows samples of the collected scores, for two videos, one from BEOID (left) and one from GTEA (right).
Annotators agree on selecting highly-relevant verbs (high score), and agree on not selecting irrelevant verbs (low scores).
These represent both result and manner verb types.
Additionally, supplementary verbs align with annotators disagreements: These are not fundamental to describing the action, but some annotators did not consider them irrelevant. 

In Fig.~\ref{fig:annotation_statistics} we present annotation statistics.
Figure~\ref{fig:annotation_statistics}(a) shows the frequency of annotations for each verb in each of the datasets. We observe a long tail distribution with the top-5 most commonly chosen verbs being \qword{move}, \qword{touch}, \qword{hold}, \qword{pick up} and \qword{pull out}.

Figure~\ref{fig:annotation_statistics}(b) sorts the verbs by their median soft assignment score over videos. We see that \qword{crack} has the highest agreement between annotators when present, whereas a generic verb such as \qword{move} is more commonly seen as supplementary. 

From these annotations, we calculate the representations in Sec.~\ref{subsec:action_representations}.
For SAMV, the soft score annotations are kept as is.
For MV, the scores are binarised with a threshold of $0.5$. Finally, SV is assigned by the majority vote.

%-------------------------------------------------------------------------

\subsection{Learning Visual Actions} \label{subsec:method}

For each of the labelling approaches described in Sec.~\ref{subsec:action_representations}, we wish to learn a function, ${\phi : \mathcal{W} \mapsto \mathbb{R}^{|V|}}$ which maps a video representation $\mathcal{W}$ onto verb labels.
For brevity, we define $\bm{\hat{y}_i} = \phi(x_i)$, where $\hat{y}_{i,j}$ is the predicted value for verb $v_j$ of video $x_i$ and $y_{i,j}$ is the corresponding ground truth value for verb $v_j$ of video $x_i$.
Typically, single label (SL) representations are learned using a cross entropy loss over the softmax function $\sigma$ which we use for both SV and VN:
\begin{equation}
	L_{SL} = -\sum_i \bm{y_i} \log(\sigma(\bm{\hat{y}_i}))
    \label{eq:crossEntropy}
\end{equation}

For our proposed multi-label (ML) representations, both MV and SAMV, we use the sigmoid binary cross entropy loss as in ~\cite{nam2014large,Rai2012simultaneously}, where $S$ is the sigmoid function:
\begin{equation}
	L_{ML} = - \sum_i \sum_j y_{i,j}\log(S(\hat{y}_{i,j})) + (1 - y_{i,j})\log(1 - S(\hat{y}_{i,j}))
    \label{eq:sigmoidCE}
\end{equation}
This learns SAMV as a regression problem, without any independence assumptions.
We note that due to SAMV having continuous values within the range [0,1], Eq.~\ref{eq:sigmoidCE} will be non-zero when $y_{i,j}=\hat{y}_{i,j}$, however the gradient will be zero.
We consciously avoid a ranking loss as it only learns a relative order and does not attempt to approximate the representation.

%%%%%%%%%%%%%%%%%%%%%%%%%%%%%%%%%%%%%%%%%%%%%%%%%%%%%%%%%%%%%%%%%%%%%%%%%%
\section{Experiments and Results} \label{sec:experiments}

We present results on the three datasets, annotated in Sec.~\ref{subsec:annotation_proc_stat}, using stratified 5-fold cross validation.
We train/test on each dataset separately, but also include qualitative results of cross-dataset retrieval.
The results are structured to answer the following questions:  How do the labelling representations compare for (i) action recognition, (ii) video-to-text and (iii)~text-to-video retrieval tasks? and (iv) Can the labels be used for cross-dataset retrieval? 

\myparagraph{Implementation}
We implemented  $\phi$ as a two stream fusion CNN from~\cite{feichtenhofer2016convolutional} that uses VGG-16 networks, pre-trained on UCF101~\cite{soomro2012ucf101}.
The number of neurons in the last layer was set to $|V| = 90$. 
Each network was trained for 100 epochs.
We set a fixed learning rate for spatial, $10^{-3}$, and used a variable learning rate for temporal and fusion networks in the range of $10^{-2}$ to $10^{-6}$
depending on the training epoch. For the spatial network a dropout ratio of 0.85 was used for the first two fully-connected layers.
We fused temporal into the spatial network after ReLU5 using convolution for the spatial fusion as well as both 3D convolution and 3D pooling for the spatio-temporal fusion. We propagated back to the fusion layer.
Other hyperparameters are the same as in~\cite{feichtenhofer2016convolutional}.

%-------------------------------------------------------------------------

\subsection{Action Recognition Results} \label{subsec:recognition}

We first present results on action recognition, comparing the standard verb-noun labels, as well as single verb labels to our proposed multiple verb-only labels.  

\myparagraph{Evaluation Metric} 
We compare standard single-label accuracy to multi-label accuracy as follows:
Let $V_i^L$ to be the set of ground-truth verbs for video $x_i$ using the labelling representations $L$, and $\hat{V}_i^L$ the top-k predicted verbs using the same representations, where $k = |V_i^L|$. For single-labels ${L \in \{SV, VN \}}$, $k = 1$, however for multi-labels, $k$ differs per video, based on the annotations. The accuracy is then evaluated as: 

\vspace*{-6pt}
\begin{equation}
	A(L) = \frac{1}{|X|}\sum_i \frac{|V_i^L \cap \hat{V}_i^L|}{|V_i^L|}
    \label{eq:vtk2}
\end{equation}
Where $|X|$ is the total number of videos. $A(\{SV, VN\})$ is the same as reporting top-1 accuracy.
For $L\in\{MV,SAMV\}$, given a video with 3 relevant verbs, if the model is able to predict two correct verbs in top-3 predictions then the video will have an accuracy of $0.66=\frac{2}{3}$. This allows us to compare the recognition accuracy of all models.
For SAMV, we threshold the annotations at 0.3 assignment score (denoted by $\alpha$), and consider these as $V_i^{SAMV}$.

\begin{table}[t]
\centering
\footnotesize
\begin{tabularx}{\textwidth}{lXXXXXXXXXXXXXX}
\toprule
    & \multicolumn{4}{c}{BEOID}      &  & \multicolumn{4}{c}{CMU}                  & & \multicolumn{4}{c}{GTEA+}        \\
    \cmidrule{2-5} \cmidrule{7-10} \cmidrule{12-15} 
    & \pphi{SV} & \pphi{VN}  & \pphi{MV}            & \pphi{SAMV}  & & \pphi{SV}  & \pphi{VN} & \pphi{MV}            & \pphi{SAMV}           & & \pphi{SV} & \pphi{VN}  & \pphi{MV}            & \pphi{SAMV}  \\ \midrule
     No. of verbs$^*$ &20 &40 &42 &90  & &12 &29 &31& 89  & &15 &63 &34 &90 \\
Accuracy & 78.1 &\textbf{93.5} & \textbf{93.0} & 87.8  & & 59.2 & \textbf{76.0} & 74.1 & 73.5  & & 59.2 & 61.2 & 71.9 & \textbf{72.9}  \\
\bottomrule
\end{tabularx}
\vspace{2pt}
\caption{Action recognition accuracy results (Eq.~\ref{eq:vtk2}). \footnotesize{$^*$for $\phi_{VN}$: number of classes in the public dataset.}} 
\label{tab:class_results}
\end{table}

\begin{figure*}[t]
  \begin{center}
  	\includegraphics[width=1\textwidth]{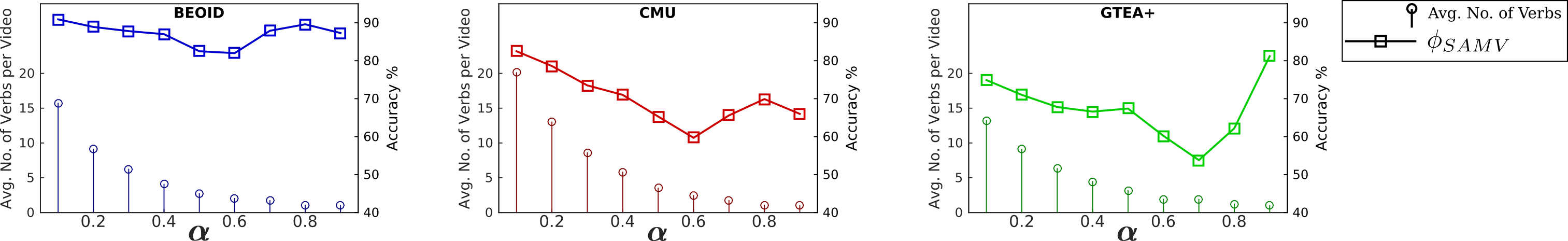}
  \end{center}
  \vspace{-20pt}
  \caption{Action recognition accuracy of \pphi{SAMV} varying $\alpha$ with avg. no. of verbs per video.} 
  \label{fig:results}
\end{figure*}

\myparagraph{Results}
Table~\ref{tab:class_results} shows the results of the four representations for action recognition. We also show the number of classes in each case.
Over all three datasets we see that $\phi_{MV}$ performs comparatively to $\phi_{VN}$ on BEOID ($-0.5\%$), worse on CMU ($-1.9\%$) and significantly better on the largest dataset GTEA+ ($+10.5\%$) due to higher overlap in actions during the dataset collection i.e. the same action is applied to multiple objects and vice versa.
~$\phi_{SV}$ is consistently worse than all other approaches reinforcing the ambiguity of single verb labels.
Additionally, $\phi_{MV}$ outperforms $\phi_{SAMV}$ on all datasets but GTEA+, where it is comparable, suggesting it is more applicable for action recognition. We note that the GTEA+ has a higher number of overlapping actions compared to the other two datasets (\ie the number of unique objects per verb is much higher).
However, $\phi_{SAMV}$ is attempting to learn a significantly larger vocabulary.

In Fig.~\ref{fig:results}, we vary the threshold $\alpha$ on the annotation scores for $\phi_{SAMV}$.
Note that for $\alpha > 0.5$, some videos in the datasets have $V_i^{SAMV} = \emptyset$ and are not counted for accuracy leading to abrupt changes.
The figure shows resilience in recognition 
suggesting the representation can correctly predict both relevant and supplementary verbs. 

%-------------------------------------------------------------------------

\subsection{Action Retrieval Results} \label{subsec:retrieval}

In this section, we consider the output of $\phi$ as an embedding space, with each verb representing one dimension.
We also evaluate whether the verb labels we have collected are consistent across datasets and provide qualitative cross-dataset retrieval results.

\vspace*{-5pt}
\myparagraph{Evaluation Metric}
We use Mean Average Precision (mAP) as defined in~\cite{philbin2007object}.

\begin{figure}[t]
  \begin{center}
     \includegraphics[width=0.9\textwidth]{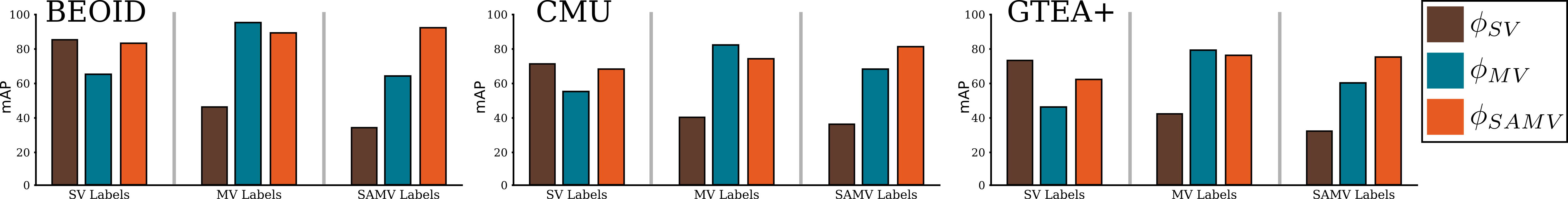}
  \end{center}
  \vspace{-12pt}
  \caption{Video-To-Text retrieval results. We show that $\phi_{SAMV}$ is able to perform well on the single verb labels and multi-verb labels but \pphi{SV} and \pphi{MV} underperform on the SAMV labels.}
  \label{fig:VT_retrieval}
\end{figure}

\vspace*{-5pt}
\myparagraph{Video-to-Text Retrieval} 
First, we evaluate $\phi$ for video-to-text retrieval. That is, given a video, can the correct verbs be retrieved and in the correct order?
Figure~\ref{fig:VT_retrieval} compares the representations on each of the labelling methods' ground truth.
Results show that $\phi_{SAMV}$ is the most generalisable, regardless of the ground-truth used,
even when not specifically trained for that ground-truth, whereas $\phi_{SV}$ and $\phi_{MV}$ only perform well on their respective ground-truth.
This shows the ability of $\phi_{SAMV}$ to learn the correct order of relevance.

\begin{figure}[t]
  \begin{center}
     \includegraphics[width=1\textwidth]{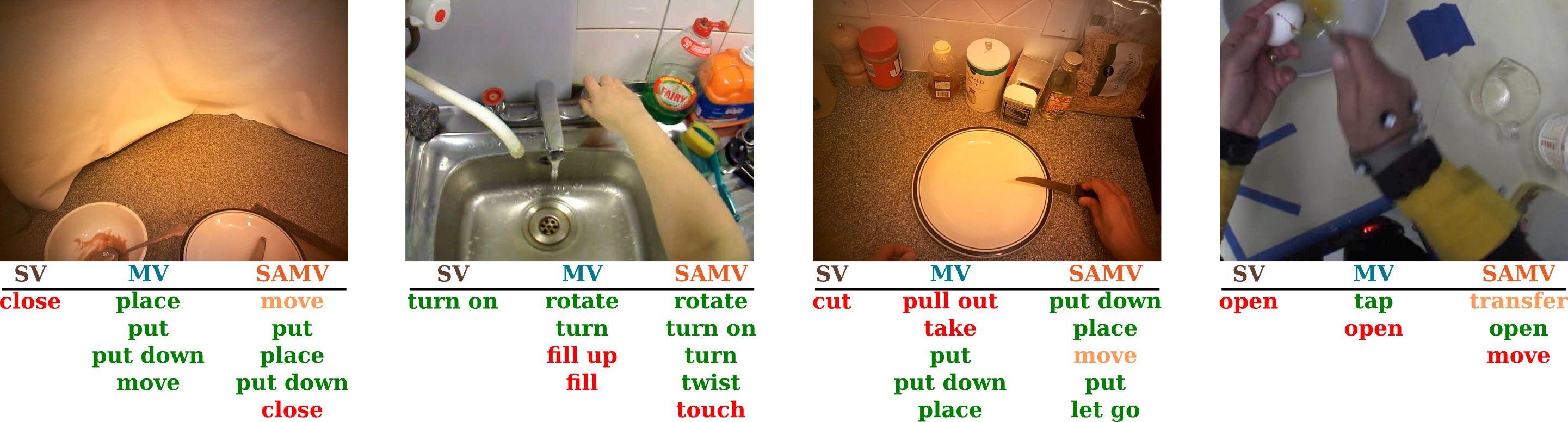}
  \end{center}
  \vspace{-10pt}
   \caption{Qualitative results comparing \pphi{\{SV, MV, SAMV\}}. Green and red denote correct and incorrect predictions. Orange shows verbs ranked significantly higher than in GT.}
\label{fig:qualitative}
\begin{center}
   \includegraphics[width=\textwidth]{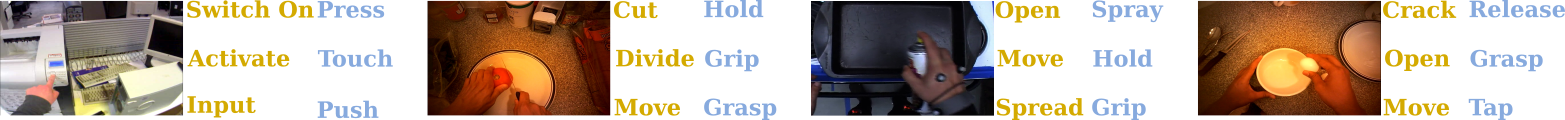}
\end{center}
  \vspace{-10pt}
   \caption{Qualitative results of the top 3 retrieved result (yellow) and manner (blue) verbs.}
\label{fig:qual_manner_result_retrieval}
\end{figure}

\begin{figure*}[b]
  \begin{center}
  	\includegraphics[width=1\textwidth]{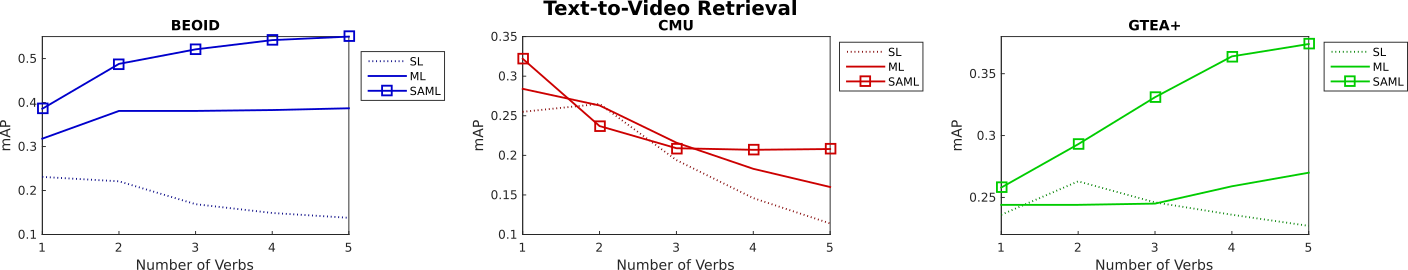}
  \end{center}
  \vspace{-10pt}
  \caption{Results of text-to-video retrieval of \pphi{\{SV, MV, SAMV\}} across all three datasets using mAP and a varying number of verbs in the query.}
  \label{fig:retrieval_results}
\end{figure*}

Figure~\ref{fig:qualitative} compares the three labelling approaches.
We also present qualitative retrieval results from \pphi{SAMV} in Fig.~\ref{fig:qual_manner_result_retrieval}, separately highlighting top-3 result and manner verbs.
We find that \pphi{SAMV} learns manner and result verbs equally well with an average root mean square error of 0.094 (for manner verbs) and 0.089 (for result verbs) respectively.

\myparagraph{Text-to-Video Retrieval} 
In the introduction we stated that a combination of multiple individually ambiguous verb labels allows for a more discriminative labelling representation.
We test this by querying the verb representations with an increasing number of verbs for text-to-video retrieval, testing all possible combinations of $n$ co-occurring verbs for $n \in [1,5]$. We show in Fig.~\ref{fig:retrieval_results} that
mAP increases for \pphi{SAMV} when the number of verbs rises for both BEOID and GTEA+ significantly outperforming other representations.
This suggests that $\phi_{SAMV}$ is better able to learn the relationships between the different verbs inherent in describing actions.
There is a drop in accuracy on CMU due to the coarser grained nature of the videos and thus a much higher overlap between supplementary verbs. 
The representation still outperforms alternatives at $n = 4$ and $n = 5$.

\begin{figure*}[t]
\begin{center}
   \includegraphics[width=1\textwidth]{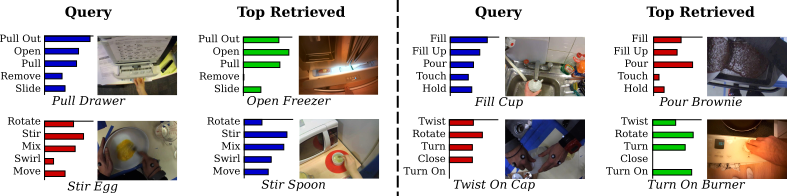}
\end{center}
  \vspace{-20pt}
   \caption{Examples of cross dataset retrieval of videos using either videos or text. Blue: BEOID, Red: CMU and Green: GTEA+.}
\label{fig:qual_vv_retrieval}
\begin{center}
   \includegraphics[width=1\textwidth]{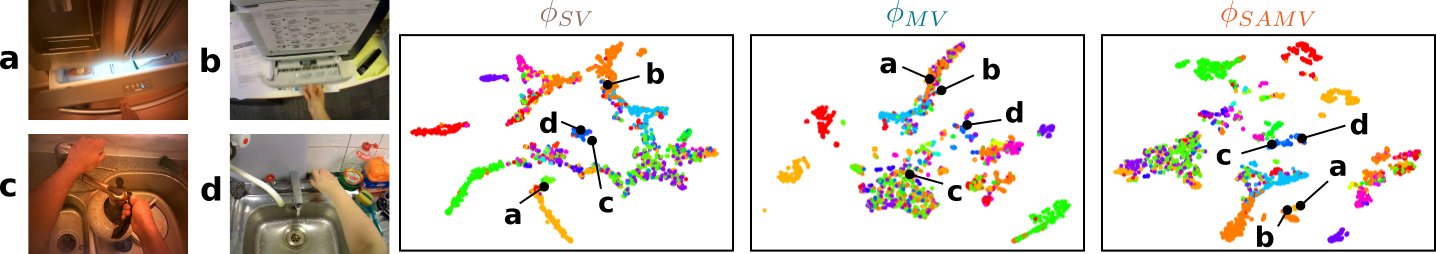}
\end{center}
  \vspace{-15pt}
   \caption{t-SNE representation of \pphi{\{SV, MV, SAMV\}} for all three datasets.}
\label{fig:qual_embedding}
\end{figure*}

\myparagraph{Cross Dataset Retrieval}
We show the benefits of a multi-verb representation across all three datasets using video-to-video retrieval (Fig.~\ref{fig:qual_vv_retrieval}).
We pair each video with its closest predicted representation from a different dataset.
\qword{Pull Drawer} and \qword{Open Freezer} show examples of the same action with different types of verbs (manner vs. result) on different objects yet their $\phi_{SAMV}$ representations are very similar.
Note the object-agnostic facet of a multi-verb representation with \qword{Stir Egg} and \qword{Stir Spoon}.

In Fig.~\ref{fig:qual_embedding} we show the t-SNE plot of the representations $\phi_{\{SV, MV, SAMV\}}$ of all three datasets.
Each dot represents a video coloured by the majority verb.
We highlight four videos in the figure.
(a) and (b) are separated in $\phi_{SV}$ due to the majority vote (\qword{pull} vs \qword{open})
but are close in $\phi_{MV}$ and $\phi_{SAMV}$.
Due to the hard assignment, (c) and (d) are far in $\phi_{MV}$ as the actions require different manners (\qword{rotate} vs. \qword{push}) but are closer in \pphi{SAMV}.

%%%%%%%%%%%%%%%%%%%%%%%%%%%%%%%%%%%%%%%%%%%%%%%%%%%%%%%%%%%%%%%%%%%%%%%%%%
\vspace*{-12pt}
\section{Conclusion} \label{sec:conclusion}
\vspace*{-6pt}

In this paper we present multi-label verb-only representations for visual actions, for both action recognition and action retrieval. 
We collect annotations for three action datasets which we use to create the multi-verb representations.
We show that such representations embrace class overlaps, dealing with the inherent ambiguity of single-verb labels, and encode both the action's \textit{manner} and its \textit{result}.
The experiments, on three datasets, highlight how a multi-verb approach with hard assignment is best suited for recognition tasks, and soft-assignment for retrieval tasks, including cross-dataset retrieval of similar visual actions.

We plan to investigate other uses of these representations for few-shot and zero-shot learning as well as further investigate the relationships between result and manner verb types, including for imitation learning.

\vspace{2pt}
\noindent \textbf{Annotations} \hspace{6pt} The annotations for all three datasets can be found at\\ \url{https://github.com/mwray/Multi-Verb-Labels}.

\vspace{2pt}
\noindent \textbf{Acknowledgement} \hspace{6pt} \dima{Research supported by EPSRC LOCATE (EP/N033779/1) and {EPSRC} Doctoral Training Partnerships (DTP). We use publicly available datasets.}
We would like to thank Diane Larlus and Gabriela Csurka for their helpful discussions.

\bibliography{epicbib}
\end{document}